\documentclass[conference]{IEEEtran}
\IEEEoverridecommandlockouts
\usepackage{cite}
\usepackage{amsmath,amssymb,amsfonts}
\usepackage{algorithm}
\usepackage{algorithmic}
\usepackage{graphicx}
\usepackage{textcomp}
\usepackage{xcolor}
\usepackage{microtype} 

\begin{document}

\title{MEND: Label-Free Detection, Localisation, and Correction of Latent Hallucination in World Models}

\author{\IEEEauthorblockN{Ali Alrasheed}
\IEEEauthorblockA{\textit{University of Melbourne}\\
Melbourne, Australia \\
ali.alrasheed@student.unimelb.edu.au}
\and
\IEEEauthorblockN{Aryan Yazdan Parast}
\IEEEauthorblockA{\textit{University of Melbourne}\\
Melbourne, Australia \\
aryan.yazdanparast@student.unimelb.edu.au}
\and
\IEEEauthorblockN{Basim Azam}
\IEEEauthorblockA{\textit{University of Melbourne}\\
Melbourne, Australia \\
basim.azam@unimelb.edu.au}
\and
\IEEEauthorblockN{James Bailey}
\IEEEauthorblockA{\textit{Monash University}\\
Melbourne, Australia \\
james.a.bailey@monash.edu}
\and
\IEEEauthorblockN{Naveed Akhtar}
\IEEEauthorblockA{\textit{University of Melbourne}\\
Melbourne, Australia \\
naveed.akhtar1@unimelb.edu.au}
}

\maketitle

\begin{abstract}
World Models are appearing as the next major frontier in computer vision. However, their robustness is currently largely unexplored. We identify the phenomenon of hallucination in latent World Models:
given a state and an action, the predicted
next latent can decode to a scene that never occurs. Because the prediction is statistically
ordinary and is fed back autoregressively by the model, the error is both silent and compounding. We study
whether such latent hallucination can be detected, localised, and corrected at inference time,
on a frozen self-supervised world model in the absence of ground-truth error labels. We introduce Masked Empirical-Bayes Neural Denoising
(MEND), a single conditional score network trained by denoising score matching on real
transitions, whose score field serves three roles: its magnitude detects
hallucination, its per-token field localises it to specific image patches, and it defines an
inference-time correction direction. On two navigation environments MEND detects hallucination
with an AUROC of up to 0.80 without using actions, exceeding a single-Gaussian density baseline
while also localising the error (per-token AUPRC up to 0.87) and correcting it, all from one
score field. Our correction reliably reduces single-step latent error and improves
predictions. We identify that a part of the error is tangent to the data manifold, hence, we focus on detection and localisation while highlighting promises of the correction.
\end{abstract}

\begin{IEEEkeywords}
world models, hallucination detection, score matching, self-supervised learning
\end{IEEEkeywords}

\section{Introduction}

\begin{figure*}[t]
\begin{center}
\includegraphics[width=0.58\linewidth]{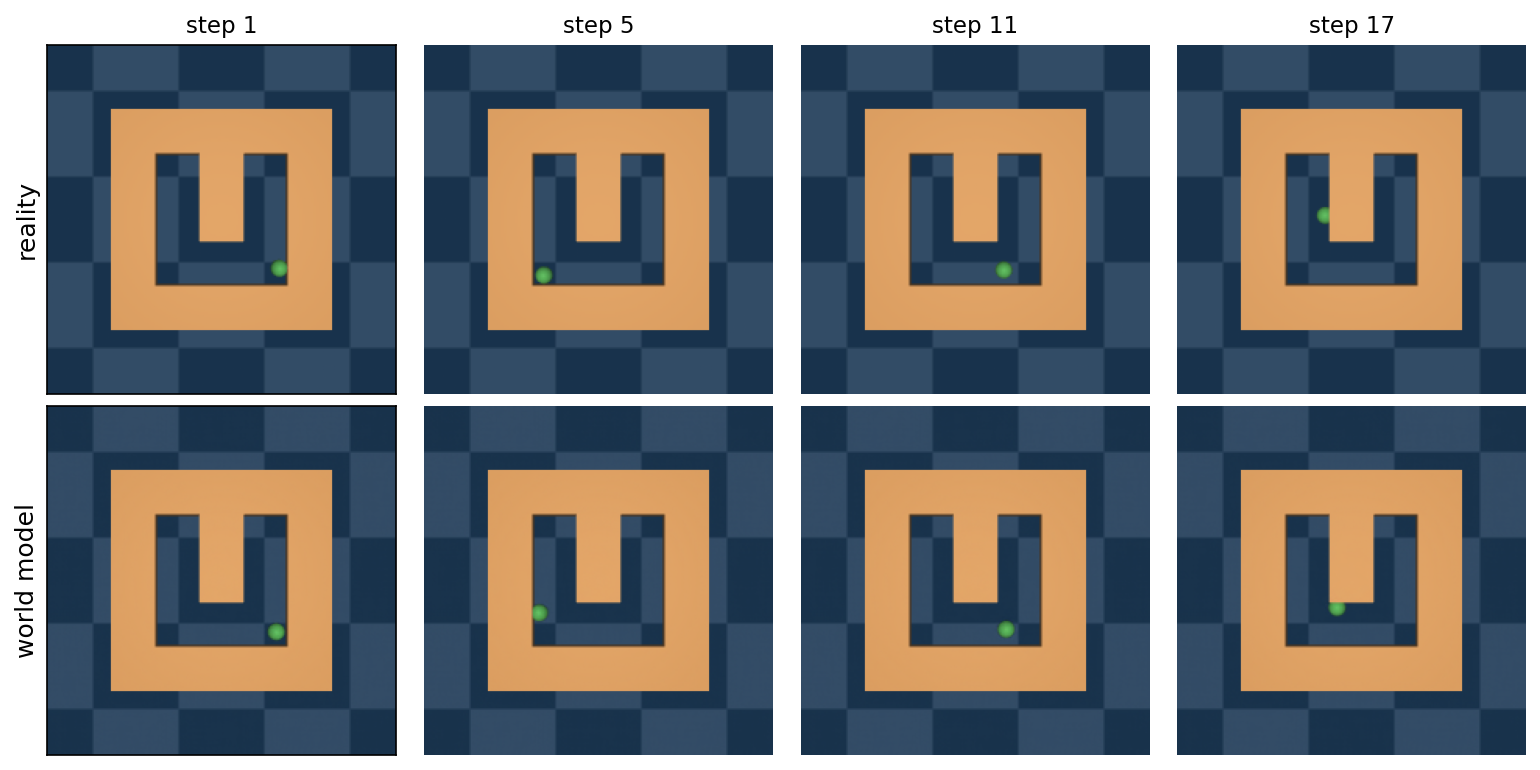}
\end{center}
\vspace{-2mm}
\caption{\textbf{Latent hallucination grows with imagination.} Free-running a frozen world
model on \textsc{PointMaze}: the top row is ground truth, the bottom row is the model's imagined
rollout decoded to images. Early on they agree, but by step 17 the imagined agent has
drifted to a part of the maze it never visits. The prediction is still a plausible
latent; nothing flags the error until the true future is known.}
\label{fig:teaser}
\vspace{-3mm}
\end{figure*}

World models are an emerging and promising approach to decision making: by imagining, or
rolling out, future latent states, they let an agent evaluate the consequences of an
action before taking it, and have been shown to solve complex downstream tasks in this
way~\cite{zhou2024dino,lecun2022path,hafner2023mastering}. Modern world models imagine future not in pixels
but in the latent space of a frozen visual predictor, which is efficient but obscure. In that
space, a model can produce a next state that is perfectly plausible as a vector of numbers
yet decodes to a scene that never happens. For instance, an object jumps across the room, a wall
dissolves, or a detail is invented where the future was uncertain. We term this phenomenon
\emph{latent hallucination}. Because the rollout is autoregressive, a hallucinated state
is fed back in as context for the next prediction. Hence, this hallucination compounds and its
effect on rollout accuracy grows with the imagination horizon (Fig.~\ref{fig:teaser}).
Because the distortion lives in the latent, it is also silent: nothing on the surface of
the prediction reveals that it is wrong, and it surfaces only once the true future
arrives, by which point a planner may already have acted on it.

Evaluation of world models has focused mainly on downstream task performance, such as
planning success or reward, and much less on whether the imagined rollout itself deviates
from reality~\cite{zhou2024dino,lecun2022path,hafner2023mastering}. Hallucination has been studied at length
for large language models, where a substantial literature now addresses how to detect and
reduce fabricated output~\cite{ji2023survey}. Similar phenomena inside world models have
received little attention, and how to detect and correct them, is largely unexplored. This
paper takes a step in that direction.

\begin{figure*}[t]
\begin{center}
\includegraphics[width=0.75\linewidth]{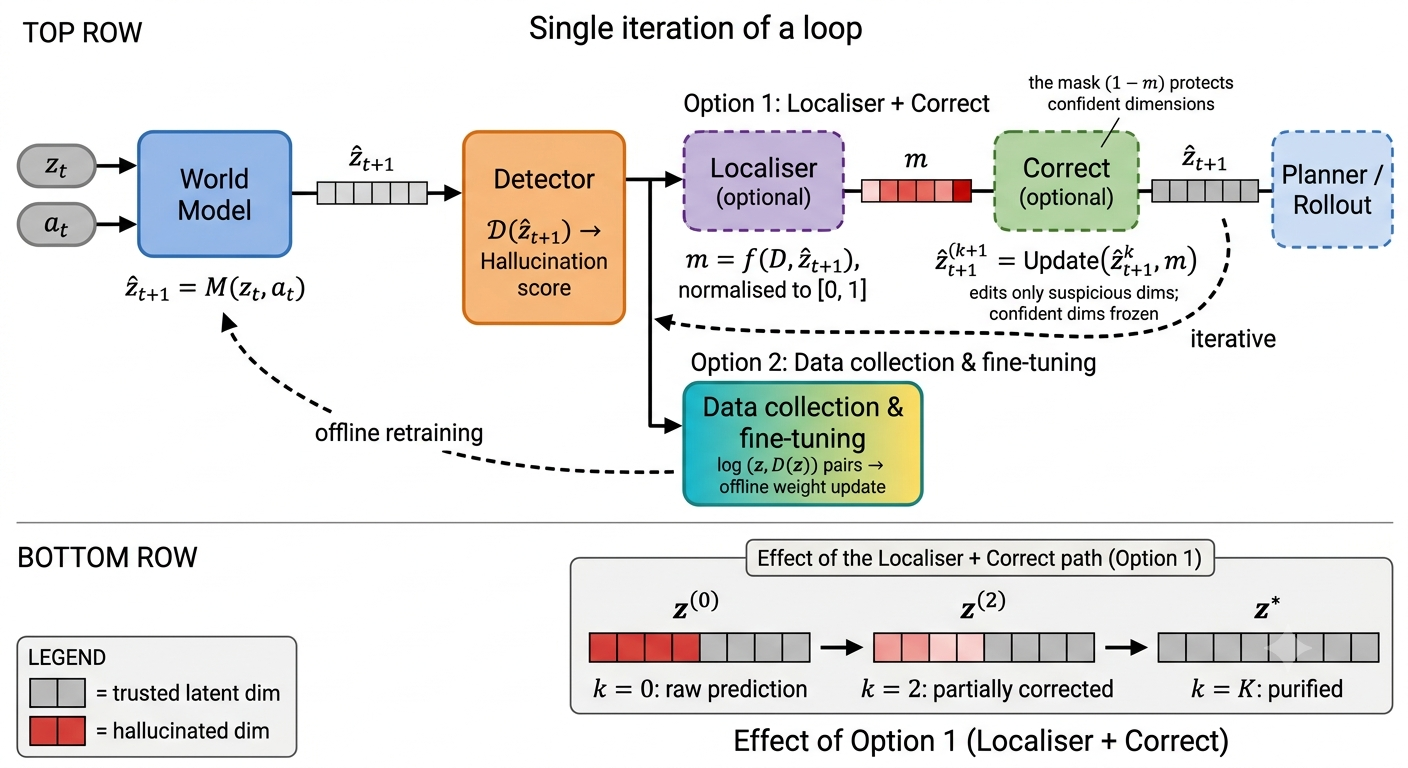}
\end{center}
\caption{MEND's single-iteration pipeline. The world model predicts $\hat{z}_{t+1}$ from
$(z_t, a_t)$ and the detector scores it, $\mathcal{D}(\hat{z}_{t+1})$. Two mitigation paths
follow. \textit{Option 1 (inference-time correction)} uses the localiser to select the
suspicious tokens ($m$) and iteratively move them toward the valid manifold while freezing the
confident tokens; \textit{Option 2 (data-driven repair)} logs the flagged predictions for
offline fine-tuning of the world model. \textbf{Bottom:} over $K$ iterations the correction
restores the hallucinated tokens (red) to trusted ones (grey).}
\label{fig:loop}
\vspace{-3mm}
\end{figure*}

Once a reliable detector exists, the most direct way to reduce hallucination is to collect
the cases where the model fails and retrain on them. This potential data-driven route is likely to be effective
but requires new data and additional training, which is generally not desirable. We instead
study an inference-time correction that operates on a frozen model, and introduce a
general way of handling hallucination in latent world models as a three-stage loop:
\emph{detect} whether a prediction is hallucinated, \emph{localise} which parts of it are
wrong, and \emph{correct} those parts (Fig.~\ref{fig:loop}). We instantiate this loop
label-free with a single conditional score field and call the resulting method \textbf{M}asked \textbf{E}mpirical-Bayes \textbf{N}eural \textbf{D}enoising (\textbf{MEND}).

MEND is a single conditional score network whose score field detects
hallucination, its per-token field localises it to image patches, and it defines an
inference-time correction direction.
We present the correction as a natural extension that complements the stronger detection
and localisation findings.
Our contributions are:
\begin{itemize}
\itemsep -2pt
\item A definition and two-type taxonomy of latent hallucination in latent world
models, showing the error is spatially sparse: a small subset of tokens carries a
disproportionate share of the total error.
\item A general detect, localise, and correct loop, instantiated label-free by
\textbf{MEND}, a single conditional score field that serves as detector, localiser, and
correcter.
\item Detection and localisation on two environments: up to 0.80 detection AUROC without
actions, exceeding a single-Gaussian density baseline while also localising the error
to specific patches (per-token AUPRC up to 0.87), and robust to the labelling threshold.
\item A promising inference-time correction loop, with analysis of what it can and
cannot recover.
\end{itemize}

\section{Related work}

\noindent\textbf{Latent world models.} A growing family of world models predicts future states not in pixels but in a learned latent space, and performs planning within that space. DINO-WM~\cite{zhou2024dino} predicts in the frozen patch-token space of DINOv2~\cite{oquab2023dinov2}; joint-embedding predictive architectures~\cite{lecun2022path}, end-to-end latent world models such as LeWM~\cite{maes2026leworldmodel}, and categorical models such as Dreamer~\cite{hafner2023mastering} follow the same paradigm. This line of work extends earlier world models that plan from pixels or recurrent latent states~\cite{ha2018recurrent,hafner2019learning} to recent diffusion-based world models~\cite{alonso2024diffusion}. Despite their architectural differences, these models rely on iterative latent prediction for planning, so prediction errors accumulate over rollout and can eventually lead to hallucination.

\vspace{1mm}
\noindent\textbf{Hallucination and uncertainty.} Hallucination is well documented for large language models and large vision language models, where a substantial body of work addresses its detection and mitigation~\cite{ji2023survey,maynez2020faithfulness,ghost}. Diffusion-based image and video generation models~\cite{ho2020denoising,rombach2022high} exhibit related failure modes, producing plausible but incorrect content. Within video generation world models, C3~\cite{mei2025world} trains a supervised probe on internal features to estimate dense per-patch confidence for video generation. In contrast, MEND targets latent world models for the first time, requiring no accuracy labels, and computing confidence directly from the predicted latent itself. The resulting confidence signal is differentiable with respect to the prediction, allowing it to be used both for hallucination detection and to guide latent correction. Detecting an implausible prediction is an instance of out-of-distribution detection, addressed with density and score-based criteria~\cite{lee2018simple,nalisnick2018deep,mahmood2020multiscale}, the density-based of which motivates our Gaussian reference~\cite{lee2018simple}. Our correction step follows score- and diffusion-based restoration, where the learned score of a data prior guides the underlying inverse problems~\cite{song2020score,kawar2022denoising,chung2022diffusion}.


\section{Definition of Latent hallucination}
\label{sec:Def}

\noindent\textbf{Setting.} In latent world models, a frozen encoder maps an observation to a latent state $z_t$,
and a frozen predictor maps a short history of states and actions to a predicted next
latent $\hat{z}_{t+1} = g(z_t, a_t)$. The latent is \emph{spatial}: it is a grid of tokens, each
describing a patch of the scene, so an error can be attributed to a region of the image.
A decoder is available to render a latent as an image, and we use it only for inspection.

\noindent\textbf{Definition.} Given the current latent state $z_t$ and action $a_t$, the
environment evolves according to its \emph{true transition} $p(\cdot \mid z_t, a_t)$: the actual
dynamics of the environment, which an ideal world model reproduces. These dynamics permit a set of
\emph{valid} next states, those consistent with the environment's physical rules, for instance
that the agent cannot pass through a wall. Let $R(z_t)$ denote the states reachable under any
action, and $R(z_t, a_t) \subseteq R(z_t)$ the subset reachable under the action taken. A
prediction $\hat{z}_{t+1}$ is \emph{hallucinated} if it lies outside this set, i.e.\ it has
negligible probability under the true transition, regardless of its prediction error. In general
these dynamics may be stochastic, so a single action can admit several valid successors; the
environments we study are deterministic, so $R(z_t, a_t)$ reduces to a single point
(Section~\ref{sec:setup}).

The same definition extends to imagined rollouts. Let $R_k(z_t, a_{t:t+k-1})$ denote the states
reachable after $k$ steps under the action sequence $a_{t:t+k-1}$. A rollout
$(\hat{z}_{t+1}, \ldots, \hat{z}_{t+H})$ is hallucination free if $\hat{z}_{t+k} \in R_k$ for
every $k$. By Markov property, its probability factorises as
\begin{equation}
\label{eq:rollout}
p(\hat{z}_{t+1:t+H} \mid z_t, a_{t:t+H-1})
= \prod_{k} p(\hat{z}_{t+k+1} \mid \hat{z}_{t+k}, a_{t+k}),
\end{equation}
where each factor is a one step transition of the form above. A single hallucinated step
therefore reduces the probability of the entire rollout and, because each prediction conditions
on previous predictions, the error accumulates with depth (Fig.~\ref{fig:teaser}). Hallucinations fall into two categories:
\begin{itemize}
\itemsep -2pt
\item[\textbf{(S)}] \textbf{Unreachable state} ($\hat{z}_{t+1} \notin R(z_t)$). A state that
cannot occur under any action, such as an object appearing from nowhere, a wall dissolving, or
the agent passing through one. It depends only on $(z_t, \hat{z}_{t+1})$.
\item[\textbf{(A)}] \textbf{Incorrect action outcome}
($\hat{z}_{t+1} \in R(z_t) \setminus R(z_t, a_t)$). The predicted state is reachable, but not
under the action that was taken. It depends on the action $a_t$.
\end{itemize}

\vspace{1mm}
\noindent\textbf{The error is sparse.} Localisation and masked correction are effective only if prediction errors are concentrated rather than uniformly distributed. In our experiments (Section~\ref{sec:setup}), this holds for both \textsc{Wall} and \textsc{PointMaze} environments. On single-step predictions, the most suspicious fifth of tokens accounts for roughly a third of the total error. During deep free-running rollouts (Fig.~\ref{fig:teaser}), the same fifth accounts for over 80 percent. Hallucination is therefore a local phenomenon that becomes increasingly concentrated as it grows, making per-token localisation and masked correction well suited to the problem.

\section{Method}

We first describe the generic detect-localise-correct loop we identify in the context of latent world models to develop a clear foundation of our method. We introduce the proposed label-free instantiation of this framework later.

\subsection{Detect, localise, and correct loop}

Once hallucinations can be detected in a rollout, they can, in principle, be mitigated in
two ways. The first is to collect the detected failures and fine-tune the world model on
them, since they identify precisely where the learned dynamics are unreliable. This
strategy directly improves the model, but it requires a data collection pipeline together
with the computational budget for retraining or fine-tuning. In many practical settings,
however, retraining is either unavailable or prohibitively expensive. In such cases, it is
desirable to reduce hallucination directly at inference time while leaving the world model
unchanged (Fig.~\ref{fig:loop}). We therefore formulate a general detect--localise--correct
loop for inference-time hallucination mitigation.

The loop begins with a detector that assigns a validity score to a predicted state. The
detector itself is intentionally left unspecified: the score may be produced by a single
model or by combining multiple models or heuristic indicators. If the same detector is
also to support localisation and correction, the score must be differentiable with respect
to the predicted state.

Differentiability allows a single score function to serve three complementary roles.
First, its value detects hallucination by measuring the validity of the prediction.
Second, its gradient with respect to the predicted state naturally localises the error:
the tokens whose perturbation would most strongly change the score are precisely those the
detector considers most suspicious, eliminating the need for a separate attribution
mechanism. Third, the same gradient provides a correction direction. By taking a small
optimisation step that reduces the detector score on only the suspicious tokens, the
prediction is iteratively refined while leaving the remainder of the latent state
unchanged.

One iteration therefore evaluates the detector, identifies the most suspicious tokens from
the score gradient, applies a masked correction step to those tokens only, and re-evaluates
the updated prediction. This process repeats until the prediction is judged sufficiently
valid or a predefined iteration budget is reached (Fig.~\ref{fig:loop}). The mask
controls the extent of the edit, trading improved error coverage against unnecessary
changes to already correct regions.

The detector, localiser, and corrector need not be separate components. Any differentiable
validity score can, in principle, provide all three functions simultaneously. In the
remainder of this section we derive such a score and use it as a unified detector,
localiser, and corrector.

\subsection{Masked Empirical-Bayes Neural Denoising}

We instantiate the loop with a single differentiable object, the score of a learned
conditional density over next states, and name the resulting method \textbf{MEND} (Masked
Empirical-Bayes Neural Denoising). The next subsections derive the score (the objective it
approximates and the network that estimates it), show how one evaluation of it fills all
three roles of the loop, and give the correction procedure in full.

\subsubsection{Mathematical basis}

A world model defines a conditional distribution over the next latent state given the current latent state and action. By Bayes' rule, this conditional probability can be expressed as - following notational conventions from above:
\begin{equation}
\label{eq:bayes}
p(\hat z_{t+1} \mid z_t, a_t) =
\frac{\;\overbrace{p(\hat z_{t+1} \mid z_t)\;
p(a_t \mid z_t, \hat z_{t+1})}^{\text{learnable}}\;}
{\underbrace{p(a_t \mid z_t)}_{\text{independent of }\hat z_{t+1}}}.
\end{equation}
Since the denominator depends only on $(z_t,a_t)$, it is constant with respect to the predicted state $\hat z_{t+1}$: it cancels when comparing candidate predictions at a fixed $(z_t,a_t)$ and disappears under differentiation. The resulting score is therefore proportional to the product of two conditional densities,
\begin{equation}
\label{eq:factors}
p(\hat z_{t+1} \mid z_t, a_t) \;\propto\;
\underbrace{p(\hat z_{t+1} \mid z_t)}_{\text{D1: reachable from }z_t}\;
\underbrace{p(a_t \mid z_t, \hat z_{t+1})}_{\text{D2: under this action}} .
\end{equation}

This decomposition forms the basis of our detector family. Each factor describes a property of the environment dynamics rather than of the world model itself, allowing it to be estimated by a small auxiliary network trained on the same logged transitions as the world model, without requiring any hallucination or accuracy labels. The resulting detectors are therefore \emph{self-supervised} and \emph{world-model-agnostic}. They assign high scores to predictions that are likely under the true environment dynamics, regardless of which world model produced them, while low scores indicate likely hallucinations.

The two factors capture complementary evidence. \textbf{D1}, the reachability term $p(\hat z_{t+1} \mid z_t)$, measures whether the predicted latent state is reachable from the current state under any valid transition, making it sensitive to type S in \S\ref{sec:Def}. \textbf{D2}, the action-consistency term $p(a_t \mid z_t,\hat z_{t+1})$, measures whether the observed action explains the transition and is therefore sensitive to type A in \S\ref{sec:Def}. Under a Gaussian action-noise model, its negative log-likelihood reduces, up to an additive constant, to the residual of an inverse dynamics model,
\[
-\log p(a_t \mid z_t,\hat z_{t+1})
\propto
\lVert a_t-\mathrm{Inv}(z_t,\hat z_{t+1})\rVert^2 .
\]

The Bayes decomposition naturally gives rise to these two complementary detector terms. Their relative effectiveness depends on the inductive biases of the learned world model and its training objective. For example, a model that primarily captures state-to-state transitions may place less emphasis on actions, making the reachability term $D1$ more informative than the action-consistency term $D2$. In our deterministic simulators, $D1$ alone provides strong hallucination detection and localisation, while $D2$ offers little additional benefit (Section~\ref{sec:detloc}). We therefore instantiate the remainder of the framework using $D1$ only, while still deriving and evaluating $D2$ as a principled ablation implied by the exact Bayes factorisation.

\subsubsection{Conditional score network}

We instantiate the framework using \textbf{D1}, the reachability density
$p(\hat z_{t+1} \mid z_t)$, and interpret hallucination as removable perturbation of a valid next
state. Specifically, we assume that a hallucinated prediction can be written as
$\hat z_{t+1} \approx z_{t+1} + e$, where $z_{t+1}$ is a valid successor of $z_t$ and $e$ denotes
the hallucination error. Rather than estimating the density itself, we estimate its
\emph{score}, the gradient of the log-density, which points in the direction of greatest
increase in validity. We therefore learn a conditional score network
$s_\theta(\tilde z \mid z_t,\sigma)\in\mathbb{R}^{N\times D}$, conditioned on the current
latent state $z_t$ and on a noise scale $\sigma>0$, the standard deviation of the Gaussian
smoothing introduced below, supplied to the network through a noise-scale embedding.

The network is trained using \emph{denoising score matching}~\cite{vincent2011connection,song2019generative}. Given
real transitions $(z_t,z_{t+1})$, we corrupt the true next state with Gaussian noise of scale
$\sigma$, $\tilde z = z_{t+1} + \sigma\varepsilon$, and train the network to predict the score of the
corrupted sample by minimizing
\begin{equation}
\label{eq:dsm}
\mathcal{L}(\theta) =
\mathbb{E}_{(z_t,z_{t+1}),\,\sigma,\,\varepsilon}\;
\sigma^2\,
\Bigl\lVert
s_\theta(z_{t+1}+\sigma\varepsilon \mid z_t,\sigma)
+\varepsilon/\sigma
\Bigr\rVert^2,
\end{equation}
where $\varepsilon\sim\mathcal{N}(0,I)$, the regression target is $-\varepsilon/\sigma$, and the
scale $\sigma$ is sampled from a geometric noise schedule. The training noise $\varepsilon$ is
distinct from the hallucination error $e$: $\varepsilon$ is an isotropic Gaussian corruption used
only during training to learn the score, whereas $e$ is the real, possibly non-Gaussian
prediction error we detect and correct at inference. As shown by~\cite{vincent2011connection}, the
population minimiser of Eq.~\eqref{eq:dsm} is the conditional score
$\nabla\log q_\sigma(\tilde z\mid z_t)$ of the $\sigma$-smoothed density of valid next
states. Denoising score matching therefore learns the score field directly, without
requiring the normalising constant of $p(\hat z_{t+1}\mid z_t)$, making estimation of D1
tractable.

This formulation has two properties that make it well suited for hallucination detection
and correction. First, the score network is trained exclusively on real transitions and
never observes predictions from the world model. It therefore learns the dynamics of the
environment rather than the behaviour of a particular predictor, preserving the
self-supervised and world-model-agnostic interpretation of Eq.~\eqref{eq:factors}. Second, although training corrupts states with Gaussian noise, that noise enters only as a
smoothing kernel and not as an assumption about the hallucination error: by
Eq.~\eqref{eq:dsm} the network learns the exact conditional score
$\nabla\log q_\sigma(\tilde z\mid z_t)$ of the smoothed density of \emph{valid} next states,
and this identity holds at every input, including a predicted state that was never formed by
adding Gaussian noise. Evaluated at a prediction $\hat z_{t+1}$, the score measures
how, and how strongly, that prediction must move to become more valid.

\subsubsection{One field, three roles}

We now make the three roles concrete. A single evaluation of $s_\theta$ at the prediction
$\hat z_{t+1}$ yields both the detection score and the localisation map; correction reuses the
same network as a short iterative refinement at a smaller noise scale
(Section~\ref{sec:setup}). The three roles therefore come from one score field rather than
three separate components (Fig.~\ref{fig:loop}).
\begin{itemize}
\itemsep -2pt
\item \textbf{Detect.} The scalar $\mathcal{D} = \lVert s_\theta(\hat z_{t+1} \mid z_t)\rVert^2$
measures how far $\hat z_{t+1}$ is from the set of reachable states. Because a regression-trained
predictor makes even its correct outputs mildly atypical, we standardise it against the
detector's statistics on known-correct predictions,
$\tilde{\mathcal{D}} = (\mathcal{D} - \mu_{\text{acc}})/\sigma_{\text{acc}}$, and use
$\tilde{\mathcal{D}}$ throughout (Section~\ref{sec:setup}).
\item \textbf{Localise.} The per-token field $\lVert s_\theta(\hat z_{t+1} \mid z_t)_n \rVert$
is a direct estimate of how far each token must move, so no attribution or backward pass
is needed.
\item \textbf{Correct.} By Tweedie's formula~\cite{efron2011tweedie} the posterior mean of the clean
state is $\hat z_{t+1} + \sigma^2 s_\theta$, so the score is also the correction direction.
\end{itemize}
The geometry behind the three roles is explicit: for a point at distance $r$ from the
reachable set, as $\sigma \to 0$ the score points from $\hat z_{t+1}$ toward its nearest valid
neighbour with magnitude $r/\sigma^2$~\cite{pidstrigach2022score}. Detection reads the length of this
vector, localisation reads its per-token parts, and correction follows it.

Correction is applied as a short iterative loop - see Algorithm~\ref{alg:mend}. MEND keeps the
original prediction $\hat z_{\text{orig}}$ as an anchor and, on the first iteration, fixes a
support of the top-$p\%$ most-displaced tokens. Each step then moves this support along the
score, a Tweedie displacement toward the valid-state manifold, while a proximal term pulls the
edit back toward $\hat z_{\text{orig}}$. The most suspicious tokens therefore move the most and the
confident ones stay near their original values. The loop returns the iterate of the lowest
standardised score $\tilde{\mathcal{D}}$. Two choices matter in practice: the correction scale
$\sigma$ must be small, since larger scales are destructive, and the correction must be
reapplied at every step of a rollout, since a single correction may not persist.

\begin{algorithm}[t]
\caption{MEND: detect, localise, correct at inference}
\label{alg:mend}
\begin{algorithmic}[1]
\REQUIRE state $z_t$, action $a_t$, predictor $g$, score net $s_\theta$, correction scale
$\sigma$, mask size $p$, step $\eta$, anchor $\rho$, budget $K$, calibration stats
$(\mu_{\text{acc}}, \sigma_{\text{acc}})$, thresholds $\tau, \delta$
\STATE $\hat z_{\text{orig}} \gets g(z_t, a_t)$; \; $z \gets \hat z_{\text{orig}}$; \;
$\text{best} \gets (z, \infty)$
\FOR{$k = 1, \dots, K$}
\STATE $G \gets s_\theta(z \mid z_t, \sigma)$ \hfill \emph{one forward pass}
\STATE $d \gets \sigma^2 G$ \hfill \emph{displacement to nearest valid state}
\STATE $\tilde{\mathcal{D}} \gets (\lVert G \rVert^2 - \mu_{\text{acc}})/\sigma_{\text{acc}}$
\hfill \emph{detect}
\STATE $m_{\text{raw}} \gets \operatorname{rank}(\lvert d \rvert)$ \hfill \emph{localise}
\IF{$k = 1$}
\STATE $S \gets \{\text{top } p\% \text{ of } m_{\text{raw}}\}$ \hfill \emph{support frozen}
\ENDIF
\STATE $m \gets m_{\text{raw}} \odot \mathbf{1}[S]$
\STATE $z \gets z + \eta\big(m \odot d - \rho\,(z - \hat z_{\text{orig}})\big)$ \hfill
\emph{correct}
\STATE \textbf{if} $\tilde{\mathcal{D}} < \text{best.}\tilde{\mathcal{D}}$ \textbf{then}
$\text{best} \gets (z, \tilde{\mathcal{D}})$
\STATE \textbf{break if} $\tilde{\mathcal{D}} < \tau$ \textbf{ or } $\lVert \Delta z \rVert < \delta$
\ENDFOR
\RETURN best iterate
\end{algorithmic}
\end{algorithm}

\section{Experimental setup}
\label{sec:setup}

\noindent\textbf{Backbone and environments.} We use the public frozen DINO-WM
checkpoints~\cite{zhou2024dino}: a DINOv2 ViT-S/14 encoder~\cite{oquab2023dinov2} with $N=196$ tokens and
$D=384$, and a small predictor. We report on two popular navigation environments, \textsc{Wall}
(a Markov predictor with a single history frame) and \textsc{PointMaze} (a three-frame
history).

\vspace{0.7mm}
\noindent\textbf{Calibration and labelling.} We hold out trajectories, run the predictor,
and split predictions into correct and incorrect at the median per-token error, giving a
balanced set. All detectors and thresholds are defined on this organic-error set. We do
not use synthetic corruptions as a headline metric, because denoising score matching is
trained to detect exactly such perturbations. The support definition of
Section~\ref{sec:Def} is not directly observable, so we label with this proxy: in the
deterministic environments studied here $R(z_t, a_t)$ is essentially a single point, so the
per-token error between a prediction and the realised next latent closely tracks whether the
prediction lies in the reachable set, and a large error means it has left that set. Under
stochastic dynamics this proxy would over-count valid alternative futures, and a
reachability-based label would be required.

\vspace{0.7mm}
\noindent\textbf{Metrics.} Detection is measured by AUROC of correct versus incorrect
predictions. Localisation is measured by per-token AUPRC against the ground-truth error
map, with a random baseline of 0.50. Correction is measured by the reduction in
latent error and the fraction of predictions improved.

\vspace{0.7mm}
\noindent\textbf{Baselines.} For detection, we compare against a single diagonal-Gaussian density
fit to valid states, the unimodal special case of feature-space Gaussian-density
detectors~\cite{lee2018simple}, which serves as a weak density floor. We further
compare an inverse-model detector (the action factor of Eq.~\eqref{eq:bayes}), a directly
conditioned score model that reads the action, and a cross-attention variant that injects the
action. Full architectures and training of these variants are given in
the supplementary material.

\noindent\textbf{Implementation.} The conditional score network is a four-layer token
Transformer (six attention heads, feed-forward width $1536$, AdaLN noise conditioning;
$10.3$M parameters, $\approx\!0.53\times$ a single world-model predictor forward), trained by
denoising score matching on roughly $500$ held-out trajectories. Detection
uses a single large noise scale $\sigma=0.39$, which probes global structure; correction uses a
small scale $\sigma=0.05$ with step $\eta=0.3$, anchor weight $\rho=0.1$, and $K=10$ Tweedie
steps ($\sigma>0.08$ is destructive).

\section{Detection and localisation results}
\label{sec:detloc}

\subsection{Across environments}

Table~\ref{tab:d1env} evaluates MEND on both environments. The central observation is that hallucination detection and localisation are related but distinct capabilities that need not improve together. Detection is a global task that asks whether an entire predicted latent state is plausible, whereas localisation is a spatial task that identifies which latent tokens are responsible for the error. Detection is strongest in environments with sharp, nearly deterministic dynamics, such as \textsc{Wall}, while localisation remains reliable whenever hallucination is concentrated to a subset of tokens, which is true in both environments. In \textsc{PointMaze}, the agent occupies only a small region of the scene and its future position is inherently uncertain, reducing the separation between valid and hallucinated predictions and lowering detection AUROC. Even so, the learned score field continues to localise hallucinations accurately (per-token AUPRC $0.87$), because the error remains concentrated in a small subset of tokens.

For detection, MEND substantially outperforms the single-Gaussian density floor on \textsc{Wall} and improves on it in \textsc{PointMaze}, demonstrating that hallucination detection requires modelling the multimodal structure of the transition distribution rather than a single-mode density. It achieves this while remaining fully differentiable, so the same object that scores a prediction also localises and corrects it.

The distinguishing advantage of MEND is that a single differentiable object simultaneously provides all three capabilities required by our framework: a global hallucination score for detection, a dense per-token localisation map, and a correction direction for inference-time refinement. A pure detection baseline such as the single-Gaussian density offers only the first of these and therefore cannot support the unified detect--localise--correct framework. This explains the absence of baseline from the last two columns.

\begin{table}[t]
\caption{MEND (\textbf{Ours}) against a diagonal-Gaussian density
reference~\cite{lee2018simple}. Detection is AUROC; the
last two columns are detection and localisation per-token AUPRC (random $0.50$).}
\begin{center}
\setlength{\tabcolsep}{3pt}
\begin{tabular}{|l|c|c|c|c|}
\hline
 & \multicolumn{2}{c|}{Detection AUROC} & Det. & Loc. \\
Environment & \textbf{Ours} & Gauss. & AUPRC & AUPRC \\
\hline\hline
\textsc{Wall}      & \textbf{0.801} & 0.648 & 0.802 & 0.712 \\
\textsc{PointMaze} & \textbf{0.691} & 0.631 & 0.687 & {0.874} \\
\hline
\end{tabular}
\end{center}
\label{tab:d1env}
\end{table}

\vspace{1mm}
\noindent\textbf{Robustness.} Detection is stable as the labelling threshold is swept from
lenient to strict, and it
saturates after a few hundred training trajectories, so the ceiling reflects the environment
and frozen backbone rather than a shortage of data (see the supplementary material).

\subsection{Localisation examples}

\begin{figure}[t]
\begin{center}
\includegraphics[width=\linewidth]{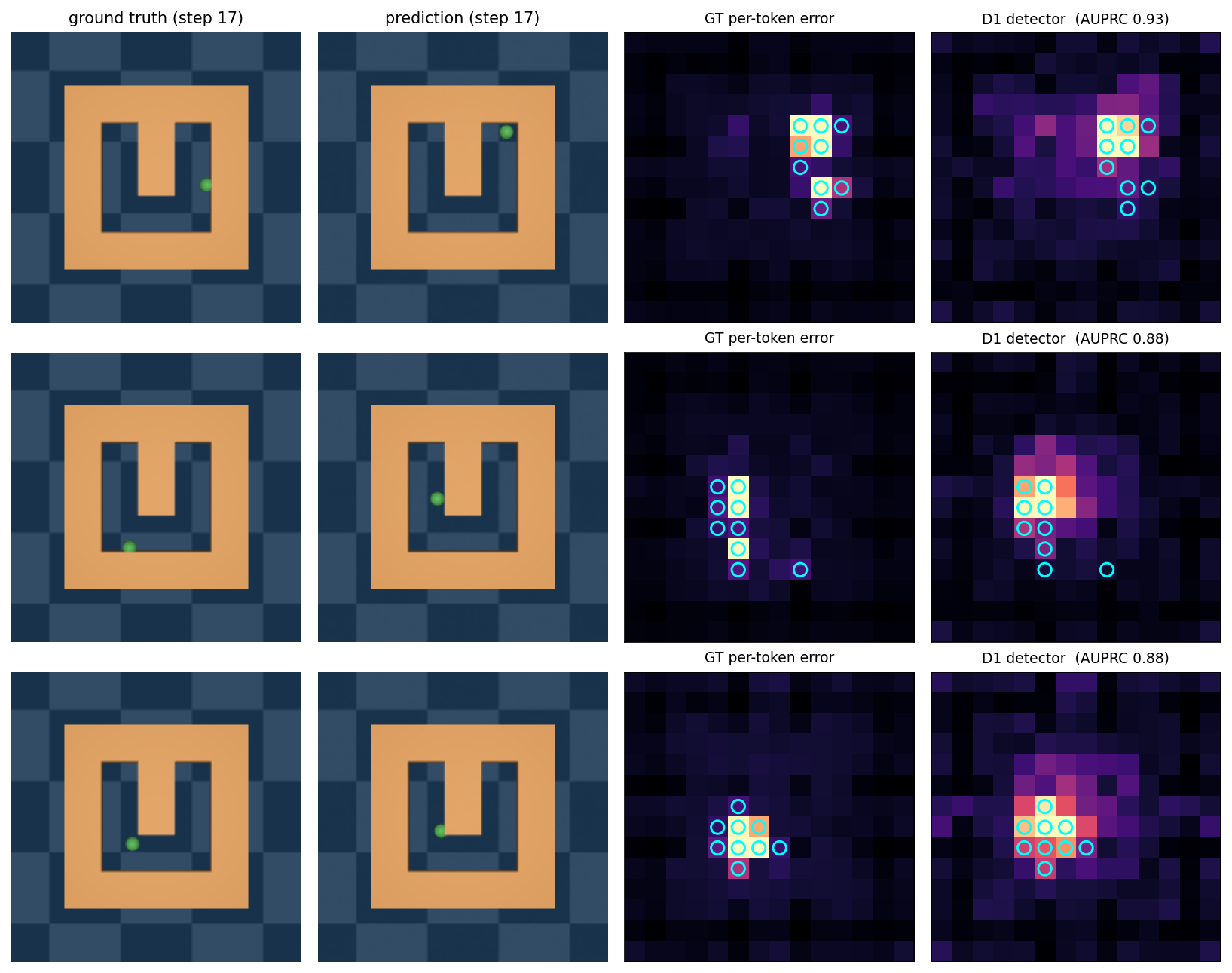}
\end{center}
\vspace{-1.5mm}
\caption{Localising \emph{deep-rollout} hallucinations on \textsc{PointMaze} (step 17 of a
free-running rollout, where the imagined agent has drifted far from reality). Columns:
decoded ground truth, decoded prediction, ground-truth per-token error, and the D1 detector
score field (per-token AUPRC annotated). Even when the whole state is far off the manifold
the detector still points at the tokens that are wrong. Cyan circles mark the nine
most-erroneous tokens.}
\label{fig:loc}
\end{figure}

Fig.~\ref{fig:loc} shows localisation on the hardest cases: hallucinations produced by
free-running the world model deep into a rollout, where the imagined agent has wandered to a
part of the maze it never visits. Each row shows the true and predicted frames, the
ground-truth per-token error, and the detector score field, which concentrates on exactly the
tokens that are wrong. The striking part is the stability with depth. As the rollout runs
forward the total error more than doubles, yet the per-token AUPRC barely moves from the
first step out to the deepest step the episodes allow. Localisation is therefore robust to
depth: even when the whole imagined state has drifted off the manifold, the same signal that
flags a hallucination keeps saying \emph{where} it is, which is what the correction stage
acts on. Equivalent deep-rollout examples on \textsc{Wall} are shown in
the supplementary material.

\section{Ablation study and discussion}
\label{sec:abl}

Throughout our experiments, the detector is D1, the reachability factor $p(\hat z_{t+1}\mid z_t)$ from the Bayes decomposition in Eq.~\eqref{eq:factors}, which depends only on the current and predicted latent states. We also evaluate the complementary action-consistency factor, D2 $=p(a_t\mid z_t,\hat z_{t+1})$, together with a variant that conditions the score network directly on the action. As shown in Table~\ref{tab:det}, neither improves hallucination detection. D2 performs at approximately chance level, while action conditioning matches, but does not exceed, the performance of D1.

This behaviour follows naturally from the deterministic setting. D2 is implemented through an inverse dynamics model, which measures how accurately the observed action can be reconstructed from the predicted transition. Because the world model is itself conditioned on the action, even hallucinated predictions generally remain consistent with that action. Consequently, the inverse dynamics residual is largely independent of whether a prediction is valid or hallucinated, making it a weak detection signal despite accurate action reconstruction.

This observation is specific to deterministic environments rather than a limitation of the Bayes decomposition itself. D2 is the only factor that distinguishes between multiple reachable future states that differ only in the action that produced them. It is therefore expected to become informative in stochastic or multimodal environments, where reachability alone is insufficient to identify the correct transition.

Conditioning the score network directly on the action likewise provides no benefit. Instead of learning the geometry of reachable latent states, the network can partially rely on the action as a shortcut. Classifier-free guidance confirms this interpretation: removing the action at training time recovers the performance of the action-free model but does not improve upon it.

\begin{table}[t]
\caption{Ablation of detector variants on \textsc{Wall}.
Variant architectures are detailed in the supplementary material.}
\begin{center}
\begin{tabular}{|l|c|}
\hline
Detector (\textsc{Wall}) & AUROC \\
\hline\hline
Conditional score net (no action) & \textbf{0.801} \\
Directly conditioned score, +action (AdaLN) & 0.799 \\
Cross-attention, +action (15M) & 0.782 \\
Inverse model --- D2 action factor & 0.490 \\
Directly conditioned score, $-$action & 0.507 \\
\hline
\end{tabular}
\end{center}
\label{tab:det}
\end{table}

\subsection{Correction results}

Correction reuses the same score field as a Tweedie step. A single such step reduces latent
error on both navigation environments and improves almost every prediction
(Table~\ref{tab:corr}). Because the Tweedie update is the per-token score field itself, each token
moves in proportion to its localisation score: the most suspicious tokens move most and confident
ones barely move, so an explicit mask is redundant and we edit the full latent. In an imagined rollout the correction must be reapplied at every step: a single first-step
correction washes out immediately, whereas per-step correction sustains a roughly constant
reduction with depth (Table~\ref{tab:rollout}; depth-resolved curves in
the supplementary material).

\begin{table}[t]
\caption{Single-step correction. $\Delta$ error is the relative change in latent error, so a
negative value indicates improvement; arrows mark the better direction.}
\begin{center}
\begin{tabular}{|l|c|c|}
\hline
Environment & $\Delta$ error $\downarrow$ & \% improved $\uparrow$ \\
\hline\hline
\textsc{Wall}      & \textbf{$-6.4\%$} & {98.5} \\
\textsc{PointMaze} & $-3.0\%$ & 99.5 \\
\hline
\end{tabular}
\end{center}
\label{tab:corr}
\end{table}

\vspace{1mm}
\noindent\textbf{What correction cannot do.} The error decomposes into a component normal to
the reachable set and a component tangent to it. Only the normal part is recoverable: a state
displaced along the reachable set is a valid alternative future, indistinguishable from the
truth for any corrector. The tangent part forms an aleatoric floor that bounds the achievable
gain, which is why the reductions are consistent but modest and why mild near-manifold errors
are corrected most. We therefore present correction as a feasibility result rather than a solved
problem.


\begin{table}[t]
\caption{Per-step correction on \textsc{Wall}: mean per-token latent error versus rollout depth
(lower is better). Parentheses show the reduction relative to No corr.}
\begin{center}
\begin{tabular}{|c|c|c|}
\hline
Depth & No corr. & MEND (D1) $\downarrow$ \\
\hline\hline
1 & 1.770 & \textbf{1.626} ($-8.2\%$) \\
5 & 2.370 & \textbf{2.237} ($-5.6\%$) \\
9 & 2.830 & \textbf{2.732} ($-3.5\%$) \\
\hline
\end{tabular}
\end{center}
\label{tab:rollout}
\end{table}

\section{Conclusion}

We showed that hallucination is not only a language-model phenomenon but also a measurable
property of latent world models, and we studied it on a frozen self-supervised backbone
without error labels. Our method, MEND, derives a single conditional score field that does three
jobs from one object: it detects hallucination, localises it to the right image patches, and
supplies an inference-time correction direction. Detection and localisation are solid and
reproducible across two environments and robust to the labelling threshold and the data budget.
Correction reliably reduces latent error but is bounded by the part of the error that lies along
the data manifold, where no corrector is likely to help. So we present it as a preliminary but promising
investigation. The same detection and localisation signal also points to a natural extension: using
it to target data collection and repair the world model directly.

In our method, detection and localisation are solid and label-free, and inference-time correction reliably
reduces latent error. However, two directions remain open. First, the approach assumes that
hallucinations lie off-manifold, which holds for the navigation environments studied
here but may not hold when a predictor's errors stay close to the manifold. Hence, extending the
study to stochastic dynamics is a natural next step. Second, correction is
established at the representation level, and whether these gains transfer to a downstream
controller is left for the future work.

\section*{Acknowledgements}

Ali Alrasheed is supported by a scholarship from Aramco. Naveed Akhtar is a recipient of the Australian Research Council Discovery Early Career Researcher Award (project \# DE230101058), funded by the Australian Government. This research was also supported by The University of Melbourne’s Research Computing Services and the Petascale Campus Initiative.

\bibliographystyle{IEEEtran}
\bibliography{egbib}

@article{zhou2024dino,
  title={{DINO-WM}: World models on pre-trained visual features enable zero-shot planning},
  author={Zhou, Gaoyue and Pan, Hengkai and LeCun, Yann and Pinto, Lerrel},
  journal={arXiv preprint arXiv:2411.04983},
  year={2024}
}

@inproceedings{
ghost,
title={{GHOST}: Hallucination-inducing image generation for multimodal {LLMs}},
author={{Yazdan Parast}, Aryan and Hosseini, Parsa and Asadollahzadeh, Hesam and {Soltani Moakhar}, Arshia and Azam, Basim and Feizi, Soheil and Akhtar, Naveed},
booktitle={The Fourteenth International Conference on Learning Representations},
year={2026},
url={https://openreview.net/forum?id=f4TACE7HhU}
}

@article{oquab2023dinov2,
  title={{DINOv2}: Learning robust visual features without supervision},
  author={Oquab, Maxime and Darcet, Timoth{\'e}e and Moutakanni, Th{\'e}o and Vo, Huy and Szafraniec, Marc and Khalidov, Vasil and Fernandez, Pierre and Haziza, Daniel and Massa, Francisco and El-Nouby, Alaaeldin and others},
  journal={arXiv preprint arXiv:2304.07193},
  year={2023}
}

@misc{lecun2022path,
  title={A path towards autonomous machine intelligence},
  author={LeCun, Yann},
  howpublished={OpenReview},
  note={Version 0.9.2},
  year={2022}
}

@article{maes2026leworldmodel,
  title={{LeWorldModel}: Stable end-to-end joint-embedding predictive architecture from pixels},
  author={Maes, Lucas and Lidec, Quentin Le and Scieur, Damien and LeCun, Yann and Balestriero, Randall},
  journal={arXiv preprint arXiv:2603.19312},
  year={2026}
}

@article{hafner2023mastering,
  title={Mastering diverse domains through world models},
  author={Hafner, Danijar and Pasukonis, Jurgis and Ba, Jimmy and Lillicrap, Timothy},
  journal={arXiv preprint arXiv:2301.04104},
  year={2023}
}

@article{ji2023survey,
  title={Survey of hallucination in natural language generation},
  author={Ji, Ziwei and Lee, Nayeon and Frieske, Rita and Yu, Tiezheng and Su, Dan and Xu, Yan and Ishii, Etsuko and Bang, Ye Jin and Madotto, Andrea and Fung, Pascale},
  journal={ACM Computing Surveys},
  volume={55},
  number={12},
  pages={1--38},
  year={2023},
  publisher={ACM New York, NY}
}

@inproceedings{maynez2020faithfulness,
  title={On faithfulness and factuality in abstractive summarization},
  author={Maynez, Joshua and Narayan, Shashi and Bohnet, Bernd and McDonald, Ryan},
  booktitle={Proceedings of the 58th Annual Meeting of the Association for Computational Linguistics},
  pages={1906--1919},
  year={2020}
}

@article{ho2020denoising,
  title={Denoising diffusion probabilistic models},
  author={Ho, Jonathan and Jain, Ajay and Abbeel, Pieter},
  journal={Advances in Neural Information Processing Systems},
  volume={33},
  pages={6840--6851},
  year={2020}
}

@inproceedings{rombach2022high,
  title={High-resolution image synthesis with latent diffusion models},
  author={Rombach, Robin and Blattmann, Andreas and Lorenz, Dominik and Esser, Patrick and Ommer, Bj{\"o}rn},
  booktitle={Proceedings of the IEEE/CVF Conference on Computer Vision and Pattern Recognition},
  pages={10684--10695},
  year={2022}
}

@article{mei2025world,
  title={World models that know when they don't know: Controllable video generation with calibrated uncertainty},
  author={Mei, Zhiting and Yin, Tenny and Baker, Micah and Shorinwa, Ola and Majumdar, Anirudha},
  journal={arXiv preprint arXiv:2512.05927},
  year={2025}
}

@article{vincent2011connection,
  title={A connection between score matching and denoising autoencoders},
  author={Vincent, Pascal},
  journal={Neural Computation},
  volume={23},
  number={7},
  pages={1661--1674},
  year={2011},
  publisher={MIT Press}
}

@article{song2019generative,
  title={Generative modeling by estimating gradients of the data distribution},
  author={Song, Yang and Ermon, Stefano},
  journal={Advances in Neural Information Processing Systems},
  volume={32},
  year={2019}
}

@article{pidstrigach2022score,
  title={Score-based generative models detect manifolds},
  author={Pidstrigach, Jakiw},
  journal={Advances in Neural Information Processing Systems},
  volume={35},
  pages={35852--35865},
  year={2022}
}

@article{efron2011tweedie,
  title={Tweedie's formula and selection bias},
  author={Efron, Bradley},
  journal={Journal of the American Statistical Association},
  volume={106},
  number={496},
  pages={1602--1614},
  year={2011},
  publisher={Taylor \& Francis}
}

@article{lee2018simple,
  title={A simple unified framework for detecting out-of-distribution samples and adversarial attacks},
  author={Lee, Kimin and Lee, Kibok and Lee, Honglak and Shin, Jinwoo},
  journal={Advances in Neural Information Processing Systems},
  volume={31},
  year={2018}
}

@article{nalisnick2018deep,
  title={Do deep generative models know what they don't know?},
  author={Nalisnick, Eric and Matsukawa, Akihiro and Teh, Yee Whye and Gorur, Dilan and Lakshminarayanan, Balaji},
  journal={arXiv preprint arXiv:1810.09136},
  year={2018}
}

@article{mahmood2020multiscale,
  title={Multiscale score matching for out-of-distribution detection},
  author={Mahmood, Ahsan and Oliva, Junier and Styner, Martin},
  journal={arXiv preprint arXiv:2010.13132},
  year={2020}
}

@article{ha2018recurrent,
  title={Recurrent world models facilitate policy evolution},
  author={Ha, David and Schmidhuber, J{\"u}rgen},
  journal={Advances in Neural Information Processing Systems},
  volume={31},
  year={2018}
}

@inproceedings{hafner2019learning,
  title={Learning latent dynamics for planning from pixels},
  author={Hafner, Danijar and Lillicrap, Timothy and Fischer, Ian and Villegas, Ruben and Ha, David and Lee, Honglak and Davidson, James},
  booktitle={International Conference on Machine Learning},
  pages={2555--2565},
  year={2019},
  organization={PMLR}
}

@article{alonso2024diffusion,
  title={Diffusion for world modeling: Visual details matter in {Atari}},
  author={Alonso, Eloi and Jelley, Adam and Micheli, Vincent and Kanervisto, Anssi and Storkey, Amos and Pearce, Tim and Fleuret, Fran{\c{c}}ois},
  journal={Advances in Neural Information Processing Systems},
  volume={37},
  pages={58757--58791},
  year={2024}
}

@article{song2020score,
  title={Score-based generative modeling through stochastic differential equations},
  author={Song, Yang and Sohl-Dickstein, Jascha and Kingma, Diederik P and Kumar, Abhishek and Ermon, Stefano and Poole, Ben},
  journal={arXiv preprint arXiv:2011.13456},
  year={2020}
}

@article{kawar2022denoising,
  title={Denoising diffusion restoration models},
  author={Kawar, Bahjat and Elad, Michael and Ermon, Stefano and Song, Jiaming},
  journal={Advances in Neural Information Processing Systems},
  volume={35},
  pages={23593--23606},
  year={2022}
}

@article{chung2022diffusion,
  title={Diffusion posterior sampling for general noisy inverse problems},
  author={Chung, Hyungjin and Kim, Jeongsol and McCann, Michael T and Klasky, Marc L and Ye, Jong Chul},
  journal={arXiv preprint arXiv:2209.14687},
  year={2022}
}

\end{document}


\title{Supplementary Material\\
MEND: Label-Free Detection, Localisation, and Correction of Latent Hallucination in World Models}

\author{\IEEEauthorblockN{Ali Alrasheed}
\IEEEauthorblockA{\textit{University of Melbourne}\\
Melbourne, Australia \\
Ali.alrasheed@student.unimelb.edu.au}
\and
\IEEEauthorblockN{Aryan Yazdan Parast}
\IEEEauthorblockA{\textit{University of Melbourne}\\
Melbourne, Australia \\
aryan.yazdanparast@student.unimelb.edu.au}
\and
\IEEEauthorblockN{Basim Azam}
\IEEEauthorblockA{\textit{University of Melbourne}\\
Melbourne, Australia \\
Basim.Azam@unimelb.edu.au}
\and
\IEEEauthorblockN{James Bailey}
\IEEEauthorblockA{\textit{Monash University \& University of Melbourne}\\
Melbourne, Australia \\
james.a.bailey@monash.edu}
\and
\IEEEauthorblockN{Naveed Akhtar}
\IEEEauthorblockA{\textit{University of Melbourne}\\
Melbourne, Australia \\
naveed.akhtar1@unimelb.edu.au}
}

\maketitle

\section{Detector variant details}
\label{app:variants}

The learned score variants below share the backbone of the main detector: a four-layer token
Transformer (dim $384$, six heads, feed-forward width $1536$) with AdaLN conditioning on the
noise scale $\sigma$ and per-token concatenation of the history window. Unless noted, they are
trained by the same denoising score-matching objective as the main detector on the same
${\sim}500$ held-out trajectories. The two non-parametric references are fit on the correct
predictions of the calibration split.

\noindent\textbf{Nearest-neighbour latent score.} A non-parametric geometric reference. Each
predicted latent is scored by its mean distance to the $k{=}10$ nearest correct next-state
latents in normalised latent space, so a large distance flags an off-manifold prediction. It
is strong because organic errors lie largely off the data manifold, which a $k$-NN search
detects directly, but it yields no localisation or correction signal.

\noindent\textbf{Diagonal-Gaussian floor.} A weak unimodal density baseline. We fit a single
diagonal-covariance Gaussian (per-dimension mean and variance) to the correct next-state
latents and score a prediction by its squared Mahalanobis distance under that covariance.
Capturing only one mode, it sits far below the geometric and learned detectors and marks the
floor of the task.

\noindent\textbf{Inverse-model detector (D2).} The action factor $p(a_t\mid z_t,\hat z_{t+1})$
of the Bayes decomposition in the main paper, realised as an inverse dynamics model
$\mathrm{Inv}(z_t,\hat z_{t+1})\!\to\!\hat a_t$. The detection score is the residual
$\lVert a_t-\mathrm{Inv}(z_t,\hat z_{t+1})\rVert^2$, which under a Gaussian action-noise model
equals $-\log p(a_t\mid z_t,\hat z_{t+1})$ up to an additive constant. $\mathrm{Inv}$ is a
small MLP over pooled features of the per-token state difference
$\Delta=\hat z_{t+1}-z_t$; because the agent occupies only a few of the $196$ patches, pooling
is weighted toward the most-changed tokens (learned attention or top-$k$) so the action signal
is not diluted. It is trained by regression to the logged action.

\noindent\textbf{Directly conditioned score (AdaLN, $+$action).} The single-model alternative
to the D1/D2 factorisation, estimating $p(\hat z_{t+1}\mid z_t,a_t)$ directly. The architecture
matches the main detector, except the action is projected to the model dim and added to the
noise-scale embedding, so every AdaLN block is modulated by both $\sigma$ and $a_t$. Setting
the action to zero recovers the action-free path.

\noindent\textbf{Cross-attention ($+$action).} A spatially selective form of action
conditioning ($15$M parameters). The action is projected into $K{=}4$ learnable action tokens,
and each Transformer block adds a cross-attention from the patch tokens to these action tokens,
while AdaLN still conditions on $\sigma$ only. Patches near the agent can therefore attend to
the action while distant patches ignore it. We optionally train it with per-sample
classifier-free-guidance dropout that zeroes the cross-attention update on a random half of the
batch, which restores the action-free behaviour used at inference and prevents the model from
relying entirely on the action (see the action-conditioning ablation in the main paper).

\section{Detection robustness}
\label{app:robustness}

\begin{figure}[t]
\begin{center}
\includegraphics[width=\linewidth]{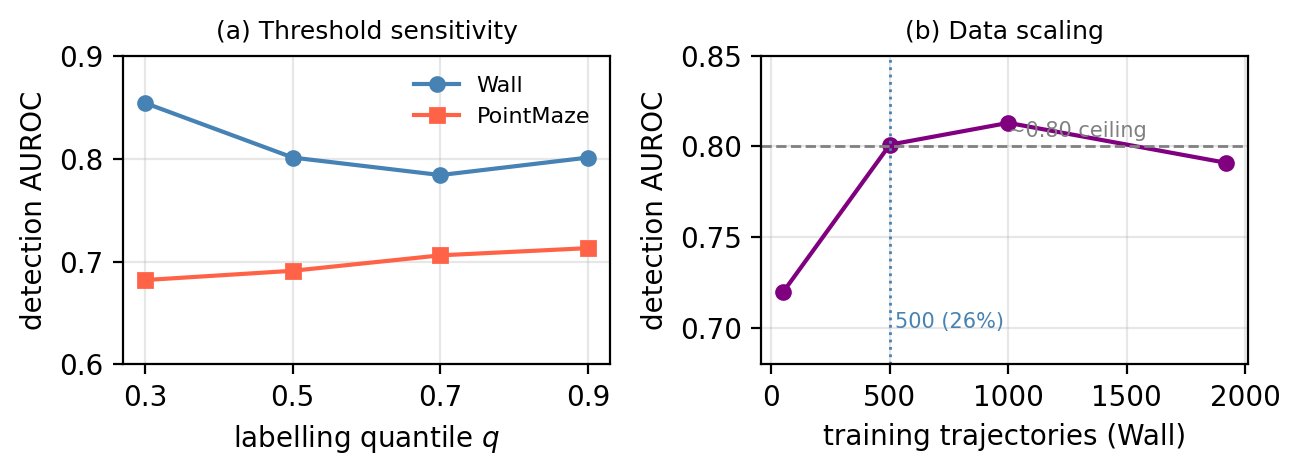}
\end{center}
\caption{Detection robustness. \textbf{(a)} AUROC is stable as the labelling quantile $q$ is
swept from lenient to strict, and improves for the worst predictions on \textsc{PointMaze}.
\textbf{(b)} On \textsc{Wall}, detection saturates near $0.80$ by ${\sim}500$ of $1920$
trajectories; more data does not help.}
\label{fig:robust}
\end{figure}

Two checks confirm that the label-free detection AUROC (D1 without actions) is not an
artefact of the labelling threshold or the training-set size (Fig.~\ref{fig:robust}).

\noindent\textbf{Labelling threshold.} We label a prediction as hallucinated by thresholding
its per-token error at a quantile $q$ of the calibration set; the headline results use the
median, $q{=}0.5$. Sweeping $q$ from lenient ($0.3$) to strict ($0.9$) leaves AUROC stable on
both environments (Fig.~\ref{fig:robust}a). On \textsc{PointMaze} the hardest predictions
($q{=}0.9$) are in fact the easiest to detect ($0.713$ vs.\ $0.691$ at the median), so the
choice $q{=}0.5$ does not inflate the reported numbers.

\noindent\textbf{Data scaling.} Detection AUROC on \textsc{Wall} rises steeply in the low-data
regime ($50\!\to\!500$ trajectories) and then flattens, saturating near $0.80$ by roughly
$500$ of the $1920$ available trajectories (Fig.~\ref{fig:robust}b). Training on the full set
does not improve detection, so the ceiling reflects a data-sufficient, right-sized model rather
than a shortage of data.

\section{Additional qualitative results}
\label{app:qual}

\begin{figure}[t]
\begin{center}
\includegraphics[width=\linewidth]{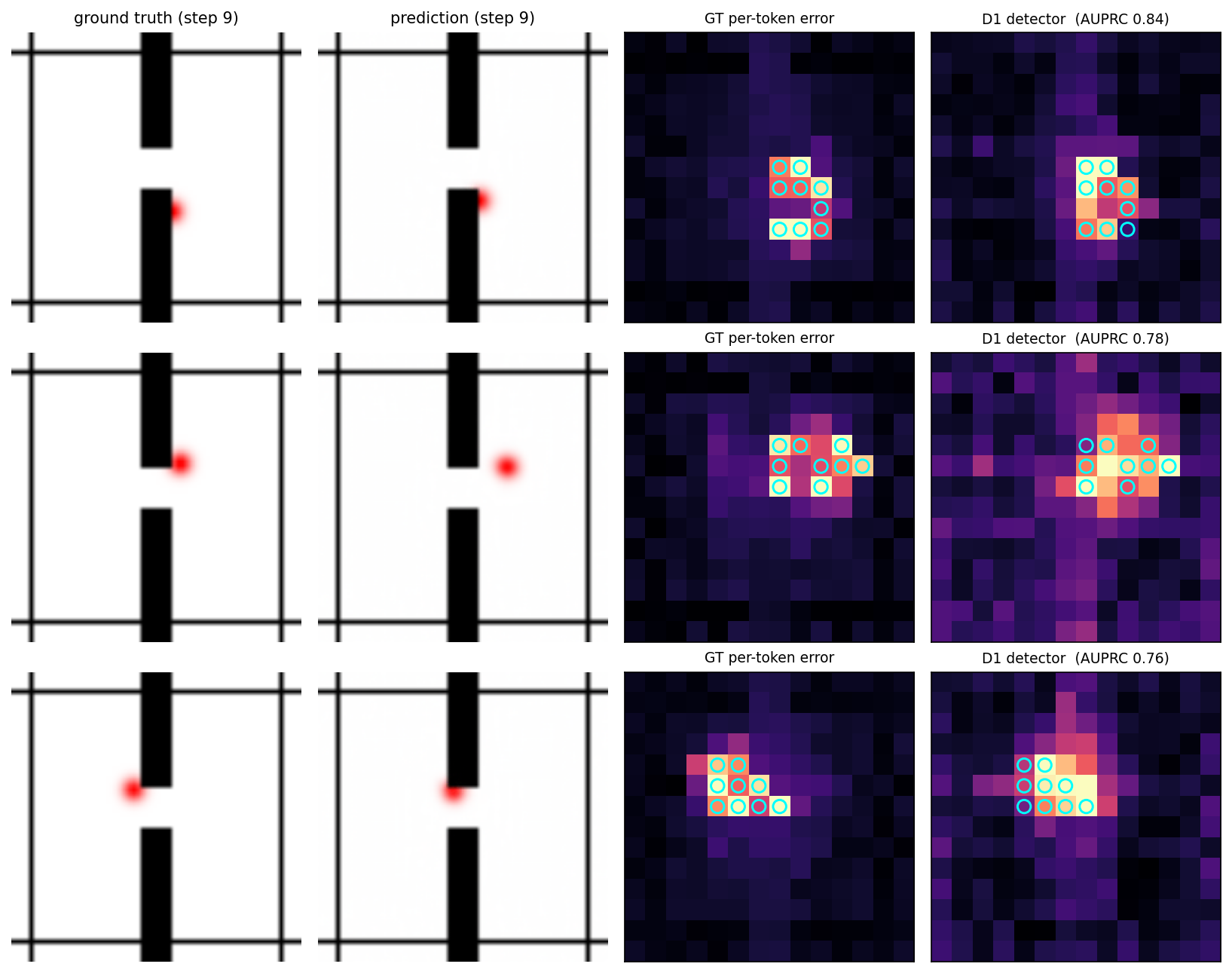}
\end{center}
\caption{Deep-rollout localisation on \textsc{Wall}. Columns: decoded ground truth, decoded
prediction, ground-truth per-token error, and the D1 detector score field (per-token AUPRC
annotated). Cyan circles mark the nine most-erroneous tokens. The score field points at the
wrong tokens even when the whole state has drifted off the manifold.}
\label{fig:loc_wall}
\end{figure}

\noindent\textbf{Localisation on \textsc{Wall}.} Fig.~\ref{fig:loc_wall} repeats the
deep-rollout localisation analysis on \textsc{Wall}, the lead environment. Each row is one
hallucination from a free-running rollout: decoded ground truth, decoded prediction,
ground-truth per-token error, and the D1 score field with its per-token AUPRC. As on
\textsc{PointMaze}, the score field concentrates on the tokens that are actually wrong.

\begin{figure}[t]
\begin{center}
\includegraphics[width=\linewidth]{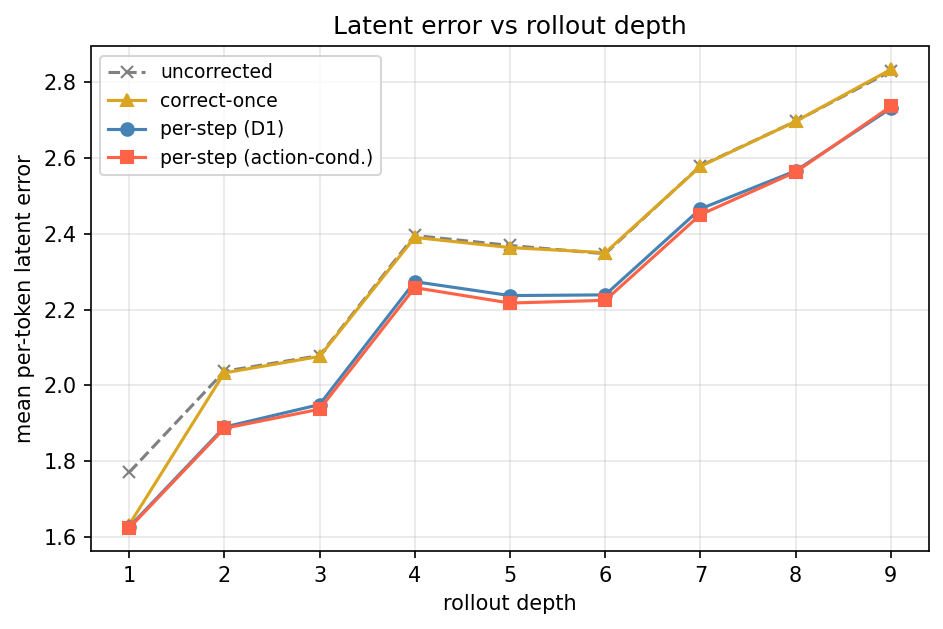}
\end{center}
\caption{Correction across an imagined rollout on \textsc{Wall}: mean per-token latent error
versus depth (lower is better). Correcting once at the first step washes out immediately, while
per-step correction sustains the reduction at every depth.}
\label{fig:corr_curve}
\end{figure}

\noindent\textbf{Correction across depth.} Fig.~\ref{fig:corr_curve} is the depth-resolved
view of the per-step correction results reported in the main paper. A single first-step
correction (correct-once) rejoins the uncorrected curve almost immediately, whereas per-step
correction (D1 and the action-conditioned variant) holds a reduction at every depth.

